# Deep learning-based denoising for fast time-resolved flame emission spectroscopy in high-pressure combustion environment

Taekeun Yoon[ab], Seon Woong Kim[ab], Hosung Byun[ab], Younsik Kim[c], Campbell D. Carter[d], and Hyungrok Do[ab]*

[a] Department of Mechanical Engineering, Seoul National University, Seoul 08826, Republic of Korea

[b] Institute of Advanced Machines and Design (IAMD), Seoul National University, Seoul 08826, Republic of Korea

[c] Department of Physics and Astronomy, Seoul National University, Seoul 08826, Republic of Korea

[d] Air Force Research Laboratory, Wright-Patterson Air Force Base, Ohio 45433, United States

* Corresponding author's Email address: hyungrok@snu.ac.kr

# Abstract

A deep learning strategy is developed for fast and accurate gas property measurements using flame emission spectroscopy (FES). Particularly, the short-gated fast FES is essential to resolve fast-evolving combustion behaviors. However, as the exposure time for capturing the flame emission spectrum gets shorter, the signal-to-noise ratio (SNR) decreases, and characteristic spectral features indicating the gas properties become relatively weaker. Then, the property estimation based on the short-gated spectrum is difficult and inaccurate. Denoising convolutional neural networks (CNN) can enhance the SNR of the short-gated spectrum. A new CNN architecture including a reversible down- and up-sampling (DU) operator and a loss function based on proper orthogonal decomposition (POD) coefficients is proposed. For training and testing the CNN, flame chemiluminescence spectra were captured from a stable methane-air flat flame using a portable spectrometer (spectral range: 250–850 nm, resolution: 0.5 nm) with varied equivalence ratio (0.8–1.2), pressure (1–10 bar), and exposure time (0.05, 0.2, 0.4, and 2 s). The long exposure (2 s) spectra were used as the ground truth when training the denoising CNN. A kriging model with POD is trained by the long-gated spectra for calibration, and then the prediction of the gas properties taking the denoised short-gated spectrum as the input: The property prediction errors of pressure and equivalence ratio were remarkably lowered in spite of the low SNR attendant with reduced exposure.

# Nomenclature

*Latin characters*

| | |
|---|---|
| $I_{dark}$ | dark current |
| $L$ | loss function |
| $N$ | number of data pairs |
| $N_c$ | number of channels |
| $N_d$ | number of sub-signals |
| $N_{electron}$ | number of counted electrons |
| $N_k$ | number of kernel size |
| $N_l$ | number of layers |
| $N_p$ | number of CNN parameter |
| $N_R$ | read noise |
| $P$ | pressure |
| $T_0$ | cosine annealing cycle step size |
| $T_{mult}$ | cosine annealing cycle magnification |
| $T_{up}$ | cosine annealing linear warmup step size |
| $W$ | number of components in input signal |
| $x_i$ | short-gated spectrum |
| $y_i$ | long-gated spectrum |
| $\hat{y}_i$ | denoised short-gated spectrum |

*Greek symbols*

| | |
|---|---|
| $\alpha$ | constant defined in Eq. (1) |
| $\gamma$ | cosine annealing decrease rate of maximum learning rate |
| $\eta$ | quantum efficiency |
| $\eta_{max}$ | cosine annealing maximum learning rate |
| $\sigma_{Photon}$ | photon noise |
| $\sigma_{Dark}$ | dark noise |
| $\sigma_{Read}$ | readout noise |
| $\tau$ | exposure time |
| $\phi$ | equivalence ratio |
| $\phi_p$ | photon flux at the CCD |

*Abbreviations*

| | |
|---|---|
| BN | batch normalization |
| CCD | charge-coupled device |
| CNN | convolutional neural network |
| Conv | convolution |
| DU | down- and up-sampling |
| FES | flame emission spectroscopy |
| HS | high-SNR |
| LS | low-SNR |
| MSE | mean square error |
| POD | proper orthogonal decomposition |
| REC | average relative errors of calibration |
| ReLU | rectified linear unit |
| REP | average relative errors of prediction |
| RSD | average relative standard deviations |
| SNR | signal-to-noise ratio |

# 1. Introduction

Recently developed fast time-resolved optical measurement methods have remarkably extended our understanding of turbulent combustion phenomena. The transient and fast-evolving combustion processes can only be investigated by non-intrusive optical diagnostics with high temporal resolution. For example, several kilohertz (kHz) sampling rates were required to temporally resolve the combustion dynamics in large-scale eddies accompanying micro-vortex structures [1-4]. However, capturing full temporal dynamics in such turbulent combustion environments is challenging even with the most advanced optical diagnostics tools. This is because various gas properties, e.g., temperature, pressure, and species concentrations, and local-bulk velocity profiles that characterize the combustion behaviors, fluctuate within extremely short characteristic time scales; typical integral and Kolmogorov time scales are less than 100 and 10 microseconds, respectively [5, 6].

Flame emission spectroscopy (FES) is the simplest optical combustion diagnostics method and therefore widely used to monitor the gas properties in the combustion zone. It was found that some spectral features, e.g., atomic emission line strength and width, molecular emission band strength, broadband spectrum profiles, etc., in the chemiluminescence photon energy spectrum are highly sensitive to the properties of the reacting medium such as equivalence ratio, pressure, and species concentration [7-10]. Therefore, calibration experiments were conducted to provide chemiluminescence spectra captured under various test conditions of accurately measured target properties, and the property indicators were calibrated for the property measurements. The calibration functions can then predict the measurement target properties by taking a chemiluminescence spectrum under an unknown condition as the input. It is noteworthy that the calibration can employ the spectra of a laminar premixed flame although the calibration functions have been used for measurements in transient turbulent flames. This is because the chemiluminescence spectrum of a turbulent flame has weak sensitivity to its strain rate and turbulent intensity and contains the same spectral peaks that behave the same as in laminar premixed flames [11-13].

In practical combustion environments, however, property variations are interconnected to affect the emission spectrum in different ways. For example, the fuel concentration increment in fuel-rich conditions will raise the hydrogen population in the medium to intensify the emission signal related to hydrogen; however, the decreased combustion temperature under the fuel-rich condition will weaken the overall signal strength at the same time. Such correlations among the flame properties reduce the measurement accuracy in conventional FES utilizing calibration functions that cannot consider the influences of multiple properties varying simultaneously in practice. Recently, it has been shown that data-driven property prediction models, e.g., partial least-squares regression, proper orthogonal decomposition with a kriging model, and neural network methods, can significantly improve prediction accuracy because these models can consider simultaneous variations of multiple properties [14-17].

FES utilizes instantaneous combustion chemiluminescence spectra for estimating the gas properties, and the exposure time for capturing the chemiluminescence spectra determines the time resolution of FES. The typical sampling rate of recently developed photon detectors is up to 10 MHz with a complementary metal-oxide-semiconductor camera [18, 19], which implies that the temporal resolution of the emission spectroscopy can significantly be improved using faster and more sensitive detectors. Recall that emission spectroscopy including FES does not require the use of any light sources while laser-aided diagnoses are limited by the repetition rate of the laser systems triggering photon emission or scattering, e.g., laser-induced fluorescence [20], coherent anti-Stokes Raman scattering [21, 22], and laser-induced breakdown spectroscopy [23].

Nevertheless, as the exposure time decreases, the signal-to-noise ratio (SNR) of the chemiluminescence spectra will decrease, with concomitant lowering of the accuracy of the FES measurements. In general, the SNR of a photon detector is proportional to the square root of the exposure time assuming negligible readout noise. Denoising filters such as Gaussian smoothing that selectively attenuate high-frequency noise [24] can enhance the SNR in cases with reduced exposure. However, the property indicators such as sharp emission peaks and radical band structures in the spectrum will inevitably be smoothed, resulting in the loss of critical high-frequency information.

The denoising ability of state-of-the-art machine learning technology for digital image processing has been investigated in numerous previous studies. Particularly, convolutional neural networks (CNN) have been extensively used for image restoration and denoising processes as a nonlinear regression method based on its architecture learning features using filters, exploiting the spatial pattern in the image dataset [25-33]. In addition, the recent development of regularization and learning methods for training deep CNN such as Batch Normalization (BN) [34], Rectified Linear Units (ReLU) [35], and residual learning [36] makes the CNN applicable to a broad range of denoising tasks. The denoising CNN (DnCNN) has been shown to achieve reasonably good performance by combining BN and residual learning, which can boost the denoising performance and accelerate training speed [26]. Moreover, a fast and flexible denoising CNN (FFDNet) can raise the denoising speed and handle a wide range of noise levels by utilizing down-sampled sub-images and an explicit noise-level map without residual learning [27]. However, these methods require noise modeling, i.e., probability distribution and noise level, which limits the application of the technique to images contaminated by real noise of unknown characteristics. The proposed denoising CNN utilizing pairs of high-quality clean and low-quality noisy data sets does not require noise modeling and performs better for eliminating the unknown noise in comparison with the conventional DnCNN [28, 29]. Recent studies such as Noise2Noise, Noise2Self, and Noise2Void, utilized only noisy data as inputs and labels to the denoising CNN [30-32]. This was enabled by assuming that the noise in the dataset is conditional independent and the zero-mean for each pixel. The denoising performances of unsupervised training with noisy data are reasonably good but the methods would not work with structured noise and clipping noise due to low-light conditions, and the denoising performance degrades quickly under high noise level conditions when not supervised by clean data [32].

These machine learning-based denoising techniques have been mostly developed for qualitative image processing rather than quantitative analyses [25-33]. The introduction of these techniques to spectroscopic analyses can make significant impacts on various quantitative optical measurement methods. Several recent studies have employed the denoising technology for spectroscopic analyses in Raman spectroscopy [37], laser absorption spectroscopy [38], electron spectroscopy [39], and angle-resolved photoemission spectroscopy [40], and it was confirmed that this machine learning technique can potentially improve the accuracy of any quantitative measurements utilizing spectroscopic analyses. It is noteworthy that CNN has also been used in other optical combustion diagnostics for nonlinear regression. Barwey et al. utilized CNN to construct velocity fields from OH planar laser-induced fluorescence images [41]. Rodriguez et al. reported that line-of-sight attenuation measurements can reconstruct the corresponding soot volume fraction field using CNN [42]. Wan et al. showed that trained CNN can distinguish combustion regime from one-dimensional Raman and Rayleigh scattering line measurements of the species mass fraction and temperature [43].

In this work, we propose a calibration process employing CNN-based denoising architecture and data mapping based on a data-driven technique to improve the temporal resolution and accuracy of FES. The denoising signal processing based on CNN is supervised by the data pairs of noisy and clean signals using a loss function that contains clean signal information quantified by proper orthogonal decomposition (POD). The proposed method removes noise based on denoising CNN architecture from the short-gated noisy chemiluminescence spectra with minimal information loss by nonlinear regression between noisy and clean data. Then, the POD method with a kriging model as a data-driven calibration technique is used for mapping the processed spectra and gas properties and evaluating prediction error. In our previous study, we found that the POD/kriging method could remarkably improve the accuracy of FES; however, the method was not very effective in handling noisy signals and thus required a denoising process for short-gated flame emission signals. It was confirmed that the proposed method combining the denoising CNN and the POD/kriging method can accurately predict gas properties in broad test ranges of equivalence ratio (0.8 – 1.2) and pressure (1 – 10 bar) of methane-air flames, taking a short-gated flame emission spectrum of low SNR as the input.

## 2. Methodology

2.1. Overall calibration and prediction process

The proposed calibration process and the test procedure of the trained model are summarized in Fig. 1. The calibration process consists of three parts: 1) Acquire training data, 2) Map data, and 3) Train CNN. Then, the prediction procedure is for measurements in arbitrary conditions: 4) Acquire high temporal resolution FES. In the following paragraph, detailed descriptions of the four parts are given.

1) Pairs of high-SNR (long-gated and averaged) and low-SNR (short-gated and instantaneous) spectra are collected in steady-state calibration experiments to train and validate the model. 2) The high-SNR spectra are used to construct a surrogate model mapping the high-SNR spectrum data to the target gas properties of interest, e.g., pressure (P) and equivalence ratio ($\phi$). Any data mapping techniques, e.g., the simplest one-to-one correlation model of emission band intensity ratios [7], partial least square regression [14], and artificial neural network [17], can be used with the proposed denoising technique. In this paper, a reduced-order model, POD/kriging is adopted because of its high accuracy and precision [16]. Property-sensitive broadband spectrum features (POD bases) are calculated by the POD that decomposes the high-SNR spectrum data, and the reduced-order model (kriging) trained by the database can accurately predict multiple gas properties taking a high-SNR spectrum as the input. More details regarding the data-mapping technique are in our previous report [16]. 3) Then, the denoising CNN is trained by the high-SNR (label) and low-SNR (input) spectrum pairs to provide a high-SNR spectrum derived from a short-gated spectrum of low-SNR (input). The hyperparameters, i.e., the parameters controlling the architecture and training process, are optimized and the neural network model is validated using a separate spectrum dataset (validation data). Finally, 4) arbitrary short-gated chemiluminescence spectrum data are collected, and high temporally resolved gas properties data are acquired.

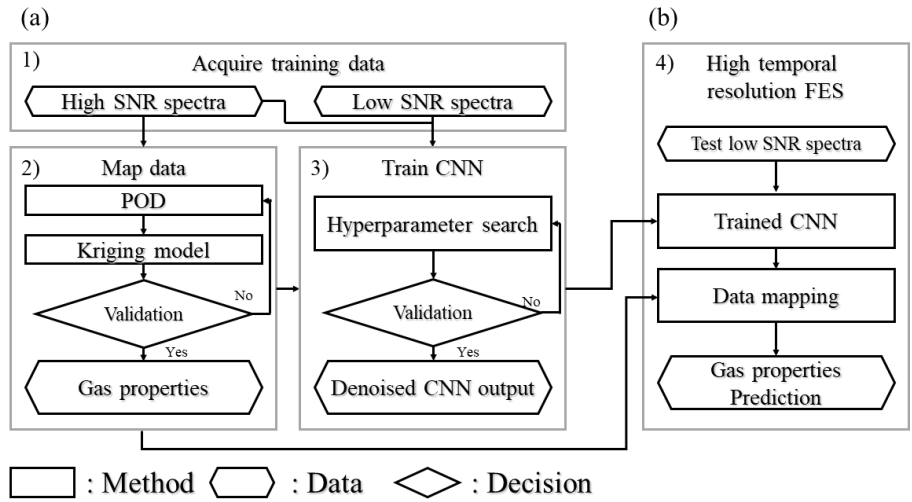

Figure 1 The sequence of (a) the calibration process and (b) the prediction procedure of the trained model.

2.2. High-pressure flat flame burner and spectrometer

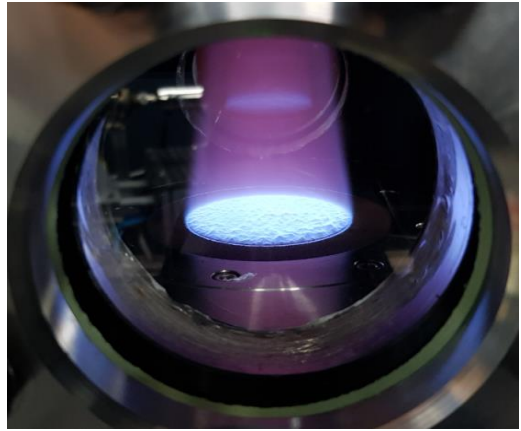

Figure 2 High-pressure flat flame burner (customized McKenna burner).

Figure 2 presents a typical flame chemiluminescence image taken on the high-pressure McKenna burner used for collecting the flame emission spectra with varied fuel concentrations and ambient pressure. The flat flame burner was installed in a high-pressure chamber that regulated the chamber pressure stably with electrically controlled mass flow controllers and a choking nozzle at the chamber exit. Fully premixed methane and air flow upward above the burner exit plane surrounded by an air co-flow. The mass flows of the gases are remotely controlled by thermal mass flow controllers (Bronkhorst EL-FLOW, F-211AV for methane, and F-211AC & F-002AV for air). The flow speed matches the laminar flame speed calculated using ANSYS Chemkin with GRI-Mech 3.0 under given gas property conditions. Sufficient diluent air, e.g., 3 times greater than the burner flow, is supplied to always keep the overall fuel concentration in the chamber below the lean flammability limit. A 60.5-mm-diameter sintered brass (porosity = 0.365) plate on the exit plane of the flat burner is water cooled so that the methane-air mixture temperature remains nearly constant.

A portable USB-connected spectrometer (Ocean optics USB 2000+, 600 grooves/mm grating) is used for collecting the flame emission spectra in a broad spectral range from 250 to 850 nm with relatively low spectral resolution (0.5 nm). A fiber-optic is connected to the entrance of the spectrometer, and a UV-camera lens (UV-Nikkor, f = 105 mm) collects flame emission and focuses the emission onto the input end of the cable. The spectra are recorded with varied exposure (gate) times, 0.05, 0.2, and 2 s. The 2 s exposure is sufficiently long to record high-SNR spectrum signals, while the reduced exposure time significantly lowers the SNR rendering characteristic spectral features less distinct. The low SNR signals are representative of FES from unstable and unsteady flames, i.e., measurement targets, where high time resolution (i.e., short exposure) is desirable.

## 2.3. Proper orthogonal decomposition of chemiluminescence signal

In this section, POD analysis on flame emission spectra is briefly introduced [16]. Figures 3 (a) and (b) present the high-SNR chemiluminescence signals with varied gas properties, $\phi$ and P. Figure 3 (a) shows that the emission peak intensities of $C_2^*$ and $CH^*$ increase as the equivalence ratio increase; Fig. 3 (b) indicates that the radical emission bands ($OH^*$, $CH^*$, and $C_2^*$) get weaker as P increases while $H_2O^*$ band is intensified. This confirms that the gas property variations affect all the critical spectral features in the flame emission spectrum differently, and their behaviors are interconnected. It was found that POD can effectively extract correlations among the spectral features and pixels of the spectrum signal that are sensitive to the properties of interest, and therefore the POD spectrum analysis could significantly improve the accuracy and precision of the property measurements using FES. Figure 3 (c) shows the property-sensitive POD bases extracted from the high-SNR training dataset, which are normalized by the average intensity of the respective $OH^*$ band (i.e., between 306 and 313 nm); the $OH^*$ band is a prominent feature in the overall gas property condition tested in this study. Basis 1 represents the P-sensitive spectrum component, Basis 2 is the most $\phi$-sensitive, and Basis 3 indicates the correlations among the emission bands. Figure 3 (d) shows that the first 4 – 5 bases contain over 99.7% of the spectrum energy, and thus any high-SNR spectrum or denoised low-SNR spectrum can be fully interpreted by combining these dominant bases (e.g., dot products of POD coefficients and POD bases) that define the relationships of all the pixels in the spectrum.

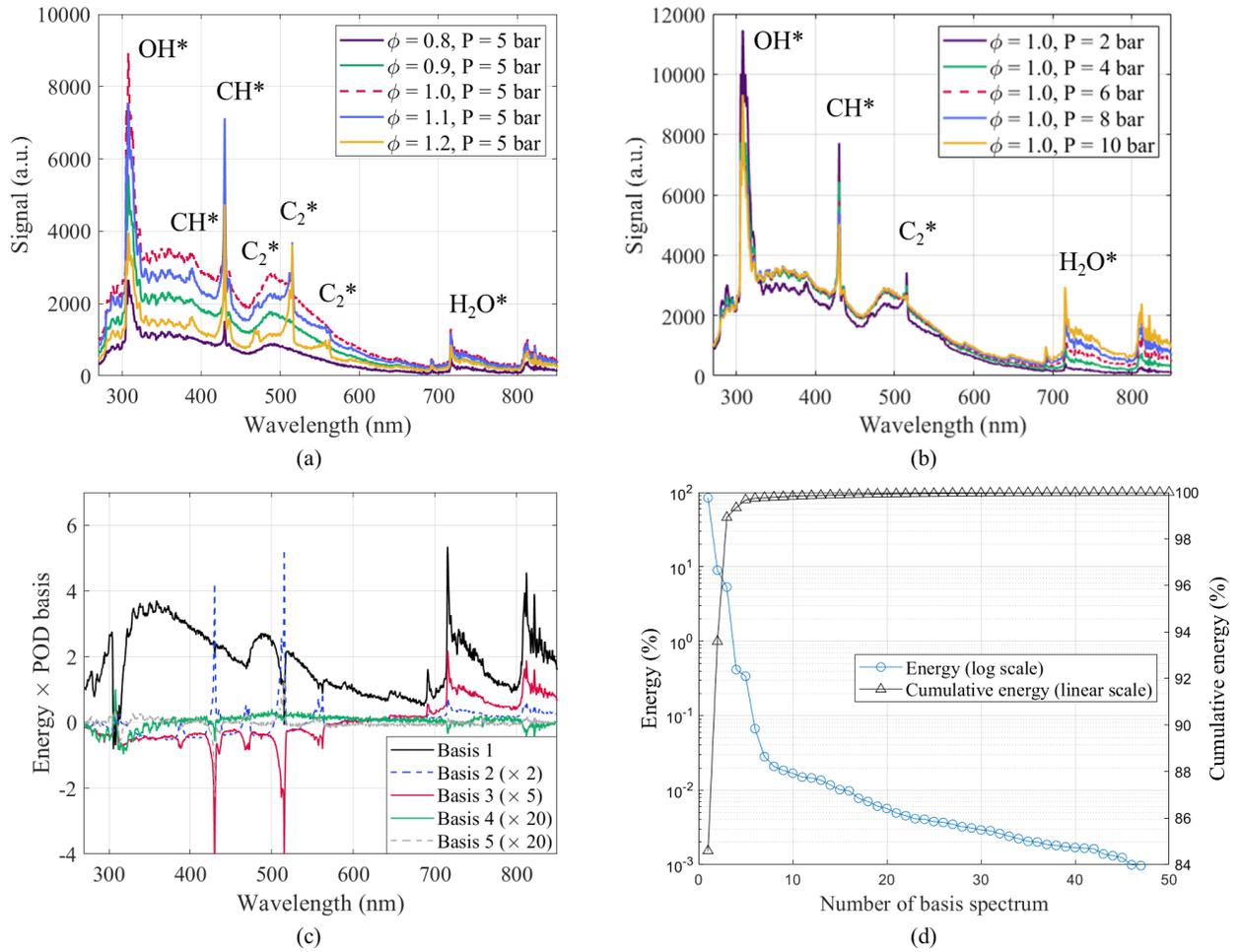

**Figure 3 (a) Chemiluminescence spectra with varied φ at P = 5bar, (b) Chemiluminescence spectra with varied P at φ = 1, (c) POD bases extracted from the training data (HS), and (d) the energy contained in primary bases.**

## 2.4. Training and test dataset

To train the neural network model, short- and long-exposure spectrum pairs $(x_i; y_i)_{i=1}^{N}$ of the chemiluminescence spectra from a stable methane-air flat flame are prepared. Here, $x_i$ and $y_i$ represent the short-gated (0.2 s) and long-gated (2 s) spectrum data, respectively. The dark spectra collected without flame are subtracted from the chemiluminescence signals. Then, the chemiluminescence signals are normalized by the respective average OH* band intensity. The total number of the data pairs used for training and testing the neural network model is 80,000: the combination of 100 low-SNR spectra and 10 high-SNR chemiluminescence signals in each of the 80 different test conditions of varied ϕ (0.8 – 1.2) and pressure (1 – 10 bar). As shown in Fig. 4, the training data denoted by green squares are sampled based on the full factorial design of 50 different gas conditions, and the test data denoted by black circles are chosen using Latin hypercube sampling of 30 gas conditions [44].

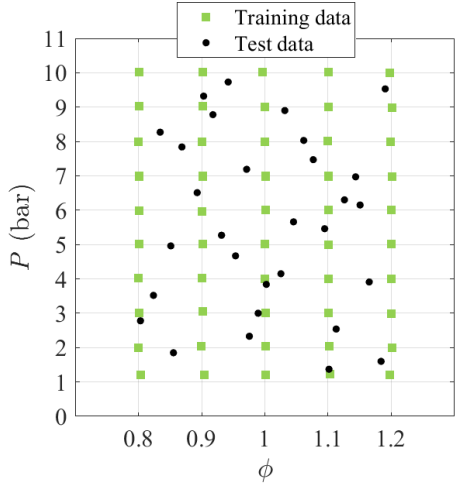

**Figure 4 Experimental data matrix.**

## 2.5. Neural network architecture

The neural network architecture for denoising the short-gated low-SNR spectra is based on FFDNet as described in Fig. 5 [27], which combines reversible down- and up-sampling (DU) operators and CNN without residual learning or skip connection. The proposed neural network architecture is supervised by pairs of high-SNR and low-SNR signals; therefore, an explicit noise-level map from a noise model is not necessary, which is different from conventional FFDNet. The input signal is a low-SNR emission spectrum recorded for a short exposure time. A reversible down-sampling operator reshaping the input signal vector of $W \times 1$ into a sub-signal tensor of $W/N_d \times N_d$, which is also known as the sub-pixel convolution or pixel unshuffle, is introduced to improve the efficiency of the neural network model. Here, $W$ is the number of components in the input signal, and $N_d$ is the down-sampling parameter determining the number of sub-signals.

The sub-signal tensor then feeds into the CNN. Each layer of the CNN is a combination of three different types of operation: Convolution (Conv), Batch Normalization (BN) [34], and Rectified Linear Units (ReLU) [35]. The first layer combines Conv + ReLU, the middle layers consist of Conv + BN + ReLU, and the last layer needs only Conv. Similar to the plain CNN architectures of Refs. [26, 27], 1) the number of channels and the kernel size of the filters are set the same for all the layers, 2) the stride (filter movement) is set to 1 without pooling to minimize data loss, and 3) the tensor is zero-padded before each CNN layer to preserve the data size. The critical hyperparameters of the CNN architecture are the depth of layer ($N_l$), the number of channels ($N_c$), and the kernel (filter) size ($N_k$).

After the last convolution layer of the last CNN layer, the reversible up-sampling operation, an inverse of the down-sampling operation, is executed to reconstruct the output signal, which has the same pixel size as the input signal. Then, the output signal is compared with the high-SNR signal (long-gated spectrum) captured at the same condition, and the performance of the model is evaluated and optimized using a loss function consisting of mean square error (MSE) and POD coefficient error. More details on the loss function are described in Sec 2.6. The hyperparameters of the neural network architecture are $N_l$, $N_c$, $N_k$, and $N_d$. A plain CNN ($N_l = 7$, $N_c = 32$, $N_k = 15$, and $N_d = 1$) as the baseline model and the DU + CNN ($N_l = 7$, $N_c = 32$, $N_k = 15$, and $N_d = 16$) are chosen to show the impact of the proposed CNN architecture. $N_c$, the number of feature extractions in CNN, is chosen as 32 considering the performance and computing cost. The performance and efficiency of the neural network model depending on the choice of the hyperparameters ($N_l$, $N_k$, and $N_d$) will be discussed later in Sec. 3.3 and 3.4.

A DU operator is utilized because the neural network architecture using a DU operator showed good performance in a wide range of noise levels by efficiently expanding the receptive field, i.e., the number of input pixels involved to produce one output pixel [27]. Expanding the receptive field can be beneficial for denoising and reconstructing signals with strong noise by utilizing information from the wide range of input signals. As shown in Fig. 6, plain CNN is compared to the proposed architecture combining plain CNN and a DU operator, which is denoted as DU + CNN. The DU operator reshapes the input signal into a set of small sub-signals that can expand the receptive field significantly without increasing network depth. The receptive field of the proposed CNN architecture is given as below:

$$\text{Receptive field} = N_d \times (N_l \times (N_k - 1) + 1) \qquad (1)$$

The receptive field can be expanded by increasing $N_l$, $N_k$, and $N_d$ with the same order of magnitude while $N_c$ does not affect the receptive field.

An ADAM optimizer is used in the optimization process to train the neural network model for 100 epochs with a batch size of 128 [45]; non-residual training is used for simplicity because the network is moderately deep, i.e., less than 20 layers [27]. A linear warmup cosine annealing learning rate scheduler is used, where cycle step size ($T_0$), cycle step magnification ($T_{mult}$), maximum learning rate ($\eta_{max}$), linear warmup step size ($T_{up}$), and decrease rate of maximum learning rate by cycle (γ) are 50, 1, 0.005, 10, and 0.1, respectively [46]. To check the overfitting of the model during the training, 10% of the training data (5 cases, 5000 data pairs) are randomly selected and set aside as the validation data. In addition, the intensity of the signal is randomly re-adjusted between 0.8 × and 1.2 × of the original signal intensity for data augmentation. The data augmentation improves the generalizability of the trained network since variations generated by the method make the training dataset diverse. We have only adjusted the signal intensity for data augmentation because flipping, rotating, and shearing the image can distort the information in the signal. The signal readjustment is a canonical and well-established method for data augmentation that improves the performance of CNN. [35, 40, 47].

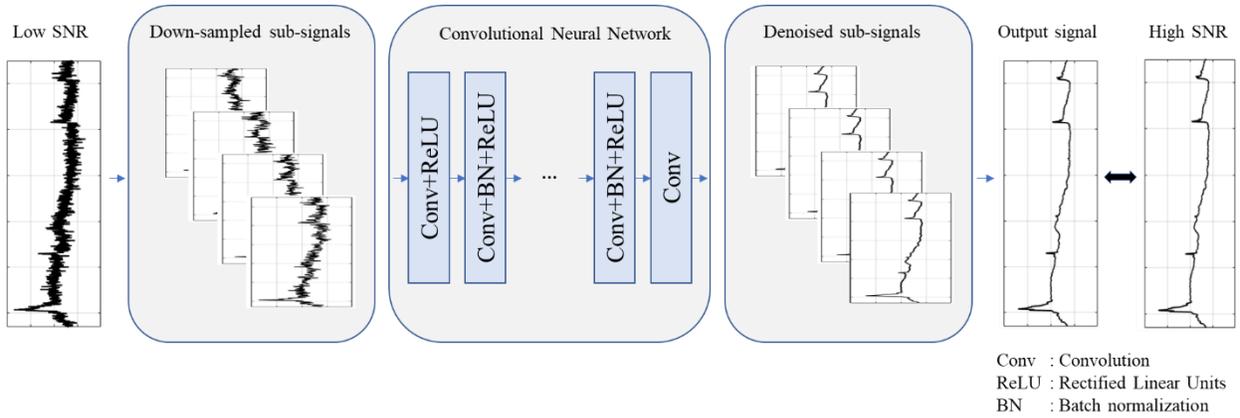

**Figure 5 A schematic diagram of the proposed denoising neural network architecture.**

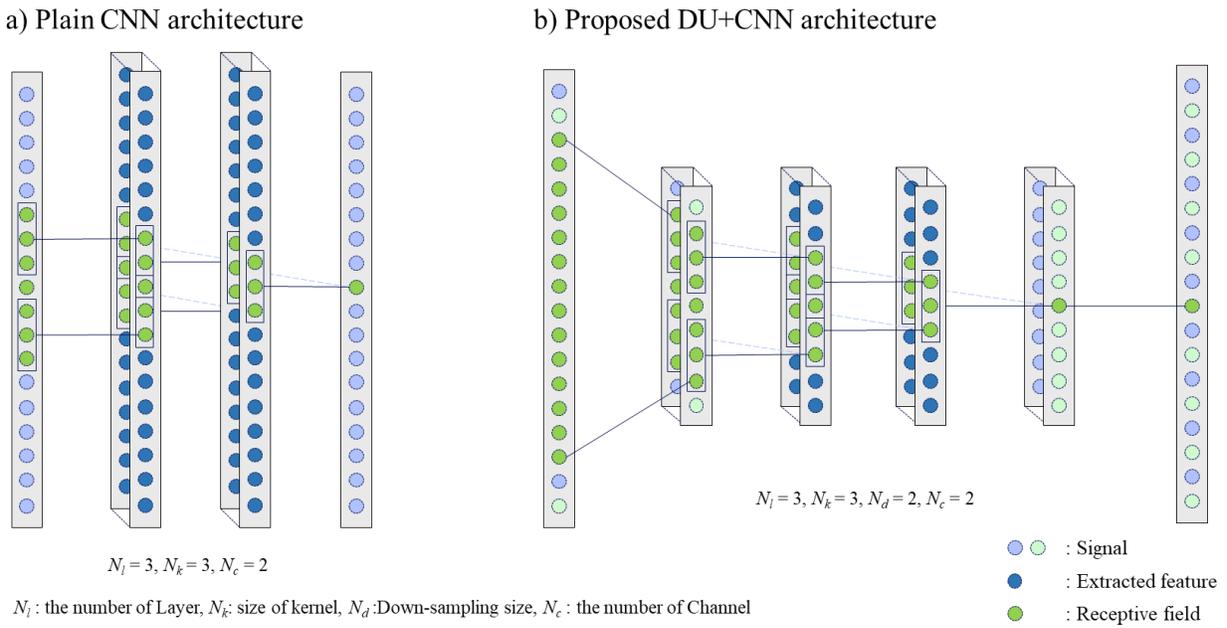

**Figure 6 Data flow of plain CNN architecture without down- and up-sampling (DU) operator and b) with DU operator.**

## 2.6. Loss function

A loss function that combines the MSE loss and a regularization term (POD loss) defined by the squared L2 norm of POD coefficients is proposed below to improve the calibration accuracy and prevent data overfitting:

$$L = \frac{1}{N}\sum_{i=1}^{N}[(1-\alpha)\|\hat{y}_i - y_i\|_2^2 + \alpha\|POD(\hat{y}_i) - POD(y_i)\|_2^2] \qquad (2)$$

where $L$, N, $\alpha$, $\hat{y}_i$, $y_i$, $POD(\hat{y}_i)$, and $POD(y_i)$ are respectively the loss function, the number of data pairs, a blending parameter, a denoised short-gated spectrum, corresponding label spectrum (high-SNR spectrum), and min-max normalized POD coefficients of $\hat{y}_i$ and $y_i$. The first term is the MSE loss frequently used for image denoising processes, and the second term is the squared L2 norm of the POD coefficient difference. The $\alpha$ is set to 0.1 to account for the magnitude scale of each loss term to be matched. Five POD bases containing 99.7% of the total energy are utilized to calculate the POD coefficient error. The $POD(x)$ operator is mapping a signal $x$ to five POD coefficients normalized by *min-max* of each coefficient of high SNR signal to set the same weight on each basis in the loss function. This is because the sensitivity to each gas property is not exactly in the order of the basis (based on energy content), whereas the variance of each POD coefficient rapidly decreases as the order increases from 1 to 5 (Fig. 3 (d)). The POD using label data can extract property-sensitive and high-dimensional (same dimension as the input spectrum) spectrum components, i.e. POD bases, from the training data set (HS). Each POD coefficient represents the weight of each POD basis to compose a spectrum (e.g., $\hat{y}_i$ and $y_i$); the dot product of the POD coefficients and the POD bases reconstructs the spectrum. The POD coefficient error, therefore, stands for the overall similarity of $\hat{y}_i$ and $y_i$, particularly, regarding the property-sensitive parts of the spectrum, which is critical information for improving the performance of the denoising CNN. This will be further discussed in Sec. 3.3.

## 2.7. Noise analysis

Noise level is quantitatively estimated by three major components assuming that the primary noise source is the charge-coupled device (CCD) sensor in the spectrometer. The major components of the CCD noise are photon noise, dark noise, and readout noise [48]. The photon noise is from stochastic photon arrival on CCD. The number of photons reaching the CCD for a fixed exposure time ($\tau$) will vary following Poisson distribution, and the square root of the error is proportional to the photon arrival. The dark noise is defined as the standard deviation of the number of spontaneously generated electrons in the CCD that also follows the Poisson distribution; the square of the dark noise, i.e., variation of generated electrons, is regarded as the number of dark electrons. The readout noise is the standard deviation of the signal level caused by the charge-voltage conversion and digitization process, which increases with faster sampling [49].

The first two noise components following Poisson distributions are approximately Normal distributions because the conditions of the central limit theorem are satisfied with a long enough exposure time. As the three components are all independent, the total error is equivalent to the root sum square of all three components. As a result, the SNR of a spectrum signal can be estimated using the equation below, Eqn. (3),

$$SNR = \frac{N_{electron}}{\sqrt{\sigma_{Photon}^2 + \sigma_{Dark}^2 + \sigma_{Read}^2}} = \frac{\phi_p \eta \tau}{\sqrt{\phi_p \eta \tau + I_{dark}\tau + N_R^2}} \quad (3)$$

where, $N_{electron}$, $\sigma_{Photon}$, $\sigma_{Dark}$, $\sigma_{Read}$, $\phi_p$, $\eta$, $\tau$, $I_{dark}$, and $N_R$ are the number of counted electrons, photon noise, dark noise, readout noise, photon flux at the CCD, quantum efficiency, exposure time, dark current, and the number of thermal electrons, respectively. Accordingly, the total error estimated predicts a monotonic decrease of SNR with decreasing exposure time.

## 3. Results and Discussion

### 3.1. Denoising with the proposed CNN

Figure 7 demonstrates the improved denoising capability of the combination of DU + CNN architecture and regularization including POD loss. The spectra used here to test the model are from the dataset that was set aside (not used for model training); this step confirms the performance of the model by taking the input of arbitrary unknown chemiluminescence signals. The chemiluminescence spectra acquired for 0.2 s and 2 s exposure are shown in Figs. 7 (a) and (d), which are denoted as low-SNR (LS) and high-SNR (HS), respectively. Characteristic local spectral features from excited molecules are clearly observed in HS spectra (Fig. 7(d)) including OH* at 306.4 nm, CH* at 431.4 nm, $C_2$* bands at 517 nm, and $H_2O$* at 700 – 850 nm. The strength and shape of the spectral features and the broadband background emission profile over the entire spectrum range are highly dependent on the gas properties in and near the combustion reaction zone and the chemical reactions in progress.

Figure 7 (b) and (c) present the outputs from a plain CNN ($N_l = 7$, $N_c = 32$, $N_d = 1$, and $\alpha = 0$) without DU operator trained by MSE loss and the proposed DU + CNN ($N_l = 7$, $N_c = 32$, $N_k = 15$, $N_d = 16$, and $\alpha = 0.1$) trained by the combination of MSE and POD loss, which are denoted as plain CNN and DU + CNN / POD loss, respectively. The two models were trained using the same sets of training and validation data. Obviously, both CNN architectures could significantly lower the noise level of the spectra compared to the short-gated noisy input signals in Fig. 7 (a). Nevertheless, the relatively weak characteristic spectral features, such as CH* at lean conditions and $H_2O$* emission bands at high-pressure conditions, that are denoised by the plain CNN model (Fig. 7 (b)) are different from those features in the corresponding HS spectra (Fig. 7 (d)). Recall that the HS spectra are the *ground truth* signals. Considering that the CH* and $C_2$* bands are highly sensitive to the fuel concentration ($\phi$) and the $H_2O$* bands are the most prominent pressure indicator as shown in Fig. 3, the gas property prediction with the plain CNN model denoising the LS spectra would be inaccurate.

On the other hand, the spectra denoised by the DU + CNN / POD loss are quite well matched with the HS spectra in most details. The noise of the input LS spectra could be successfully suppressed without missing critical spectral features under various gas property conditions, regardless of the noise level with the DU operator ($N_d = 16$). We conjecture that this is because of proper down-sampling that makes a wider receptive field, which is the number of input pixels involved to produce one output pixel, and the CNN architecture is regularized by POD loss. In other words, each pixel of the output signal is denoised and reconstructed by global spectral features of the input signal with a large receptive field and guided by the POD coefficients of each basis mode. Therefore, the successful denoising process is enabled by the model decoupling the property-sensitive spectral features from the noise signal even though the noise level changes with the randomly re-adjusted overall signal intensity. This will be further discussed in Sec. 3.3.

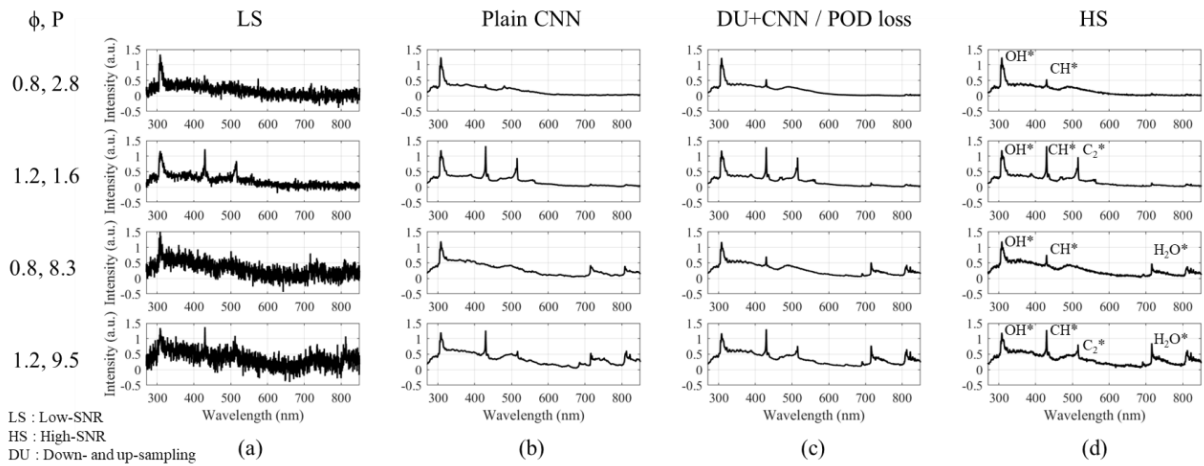

**Figure 7 Flame emission spectra captured for (a) 0.2 s (Low-SNR, LS) and (d) 2 s (High-SNR, HS) exposures, and the LS spectra denoised and reconstructed by (b) plain CNN and (c) DU + CNN / POD loss.**

## 3.2. Calibration and prediction of gas properties

To evaluate the validity of the proposed CNN quantitatively, gas properties (P and ϕ) are predicted by calibrating 1) the raw LS spectrum signals, and the emission spectra denoised-reconstructed by the 2) plain CNN and 3) DU + CNN / POD loss architectures using the same definition provided in the previous section. The x-axis in Fig. 8 denotes the ground-truth values which are measured by high-accuracy sensors, and the y-axis denotes the predictions by the short-gated (0.2 s) chemiluminescence spectra (a) without and with the denoising CNN of (b) plain CNN, and (c) DU + CNN / POD loss. The error bars indicate the standard deviations of the gas property predictions from 100 short-gated LS chemiluminescence signals, and the uncertainty bands of the sensor-measured P and ϕ are also presented in the figures (gray dotted lines), which are calculated using the uncertainty propagation equation by a first-order Taylor series expansion [50]. Even though the plain CNN could effectively denoise the LS spectrum signals as shown in Fig.

7 (b), the accuracy of the gas property prediction via calibrating the plain CNN processed spectra is not much improved compared to the prediction by the raw LS signals without the denoising CNN process. However, DU + CNN / POD loss model remarkably improves the accuracy and precision of the prediction of P and ϕ in Fig. 8 (c). This is because DU + CNN / POD loss can effectively remove the noise from the LS spectrum signals with minimal information loss utilizing the DU operator and POD loss as confirmed in Fig. 7 (c). Recall that the multiple molecular bands (OH*, CH*, $C_2$*, and $H_2O$*) in the flame emission spectrum are distributed in a broad wavelength range and are closely interconnected, as shown in Fig. 3. Therefore, the wider receptive field of down-sampled signals and regularizing training that considers the interconnections between the emission bands utilizing the POD loss can better preserve the global molecular band features that are sensitive to the gas properties; thereby, the accuracy of the gas property prediction is improved.

Table 1 summarizes the performance of property-prediction based on LS, plain CNN, and DU + CNN / POD loss. The average relative errors of calibration using the training data (REC), the relative errors of prediction with the test data (REP), and the relative standard deviations of the prediction (RSD) are tabulated, which are to evaluate the calibration accuracy of training data, the prediction accuracy of test data, and the prediction precision of test data, respectively [51]. As expected, DU + CNN / POD loss outperforms LS and plain CNN in prediction accuracy and precision. Particularly, the respective accuracy (REP) and precision (RSD) of the ϕ-prediction are only 1.5% and 1.6%, which are comparable to the sensor measurement uncertainty. Considering that the exposure of the portable spectrometer is only 0.2 seconds, the accuracy of the ϕ-prediction taking the short-gated flame emission spectrum as the sole input is exceptionally high.

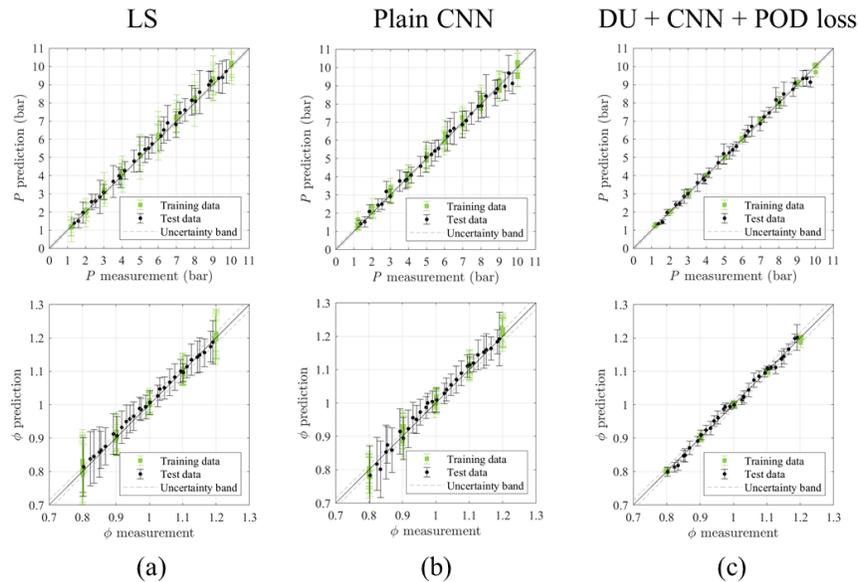

**Figure 8 Pressure (P) and equivalence ratio (ϕ)-prediction using (a) low-SNR (LS) spectra, (b) output using plain CNN, and (c) output using DU + CNN / POD loss.**

**Table 1 Performance of the three schemes**

| Error Unit: % | LS | | | Plain CNN | | | DU + CNN / POD loss | | |
|---|---|---|---|---|---|---|---|---|---|
| | REC | REP | RSD | REC | REP | RSD | REC | REP | RSD |
| $P$ | 14 | 12 | 14 | 10 | 9.9 | 12 | 2.2 | 5.7 | 6.4 |
| $\phi$ | 4.8 | 4.1 | 5.0 | 3.6 | 4.2 | 5.0 | 0.56 | 1.5 | 1.6 |

*LS: low-SNR, REC: average relative error of training data in calibration, REP: average relative error of test data in prediction, RSD: average relative standard deviation of test data in prediction.

## 3.3. Neural network architecture and loss function

Hyperparameter settings for the neural network architecture and the configuration of the loss function are important because they determine the capacity, efficiency, and performance of the model. Figure 9 illustrates the REP distribution depending on the choice of hyperparameters and loss function configurations; the REP, prediction error of P and $\phi$, represents the performance of the CNN architecture. The origin of the graph in Fig. 9 means zero prediction error for both P and $\phi$; therefore, the prediction accuracy improves as the markers get closer to the origin. The blue color of the markers in Fig. 9 denotes the prediction accuracy with the newly proposed loss function combining MSE and POD losses, and the green color indicates the REPs with the conventional loss function considering only MSE loss. The brightness of the color indicates $N_l$, which increases from 2 to 15 as it gets darker, and the shapes of the marker indicate six different combinations of $N_k$ (3, 15, and 25) and $N_d$ (1 and 16).

In general, the prediction error decreases as the $N_l$ and $N_k$ increase because of the model complexity. More importantly, it is evident that the DU operator ($N_d = 16$) and the new loss function (MSE + POD) remarkably reduce the error. On the other hand, the prediction errors of some cases with the plain CNN utilizing the conventional MSE loss function are greater than that of raw LS spectra without any denoising process. This is because the denoising CNN can misinterpret the property information contained in the raw spectrum data when misguided by the MSE loss function. Therefore, the loss function that additionally considers the POD loss should be used for denoising the short-gated LS spectra, which helps to preserve and reconstruct the property-sensitive spectral features.

Recall that a reduced-order model (kriging) predicts the gas properties based on the denoised LS input spectrum profile being decomposed by the dominant bases; the decomposition process brings out the POD coefficients (weights of the bases) to calculate the gas properties via the kriging model previously trained by the training dataset. Therefore, the new loss function including the POD loss can preserve global features of LS signal by decoupling random noise signal from property-sensitive spectrum information and prevent over-fitting, while the conventional loss functions simply consider the mean square errors (MSE) of the spectra regardless of the sensitivity of the spectrum profile to the properties of interest.

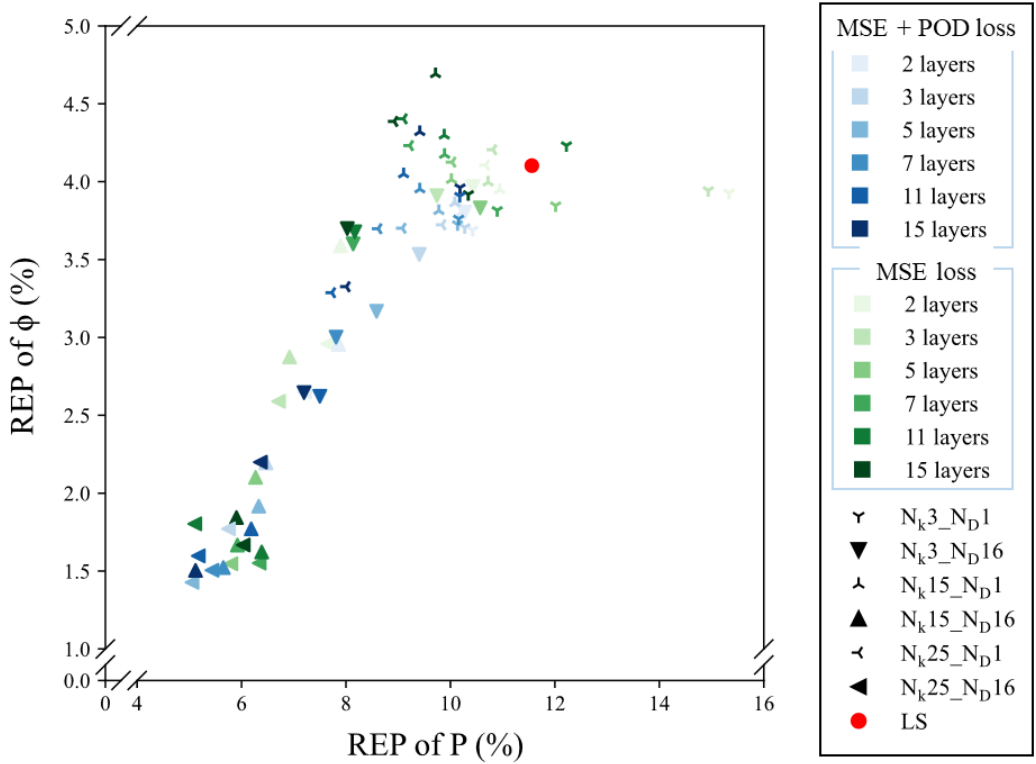

**Figure 9** Average relative errors of prediction (REP) distribution of P and φ with different combinations of two loss functions, six pairs of $N_k$ (3, 15, and 25) and $N_d$ (1 and 16), and six $N_l$ (2, 3, 5, 7, 11, and 15). Double diagonal lines on the axes indicate broken axis to show the origin.

In this study, we found that the receptive field is another important control parameter determining the performance of the deep-learning architecture developed for analyzing spectra. Figure 10 describes the non-monotonic trends of REP and RSD as functions of the receptive field. The $N_l$ (2 – 15), $N_k$ (3 – 45), and $N_d$ (1 – 32) are varied in wide ranges to reveal the impact of the receptive field on the model performance, not the separate influences of the three parameters as in Fig. 9. $N_d$ is 1 for 'plain CNN' and varied from 2 to 32 for 'DU + CNN', and both neural networks are regularized by MSE + POD loss for a fair receptive field comparison. The REP and RSD are nearly constant with the receptive field under 100 but decrease rapidly until the receptive field reaches the pixel number of the input spectrum (W, 1696 × 1), and both the performance indicators (REP and RSD) approach minima at around 2 × W. It is noteworthy that each of the output pixels will be constructed using all the input pixels when the receptive field is 2 × W; therefore, further increase of the receptive field beyond 2 × W will not help to improve the model performance. In conclusion, the receptive field of the CNN needs to be set between W and 2 × W to minimize the REP and RSD, or around W to reduce the model complexity with acceptable prediction accuracy and precision.

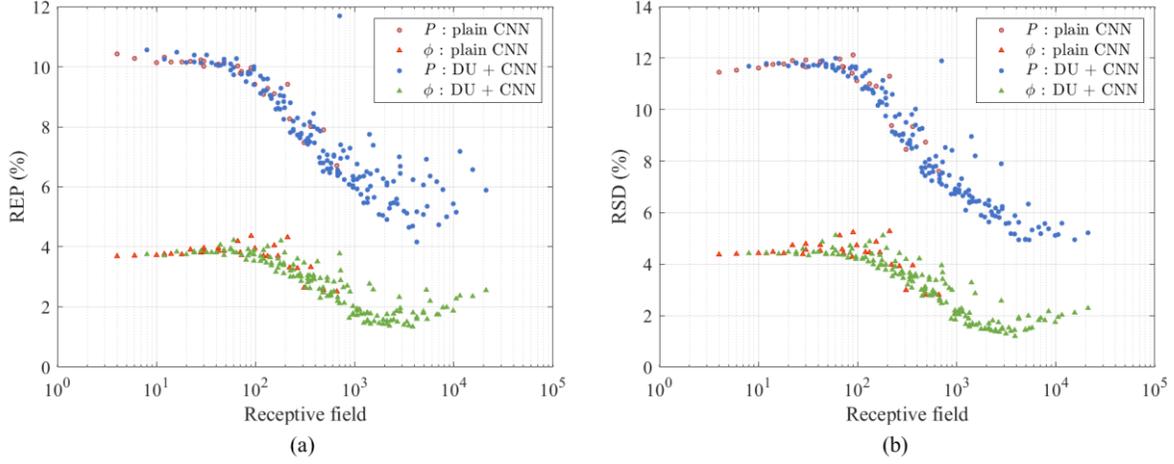

**Figure 10 (a) REP and (b) RSD of P and ϕ versus receptive fields of plain CNN ($N_d$ = 1) and DU + CNN ($N_d$ = 2 – 32) trained by POD loss**

In previous investigations on denoising CNN architecture for 2D images, the optimal receptive field size has been discussed, e.g., between 35 × 35 and 61 × 61 [27]. However, the flame emission spectrum has a unique characteristic clearly distinguished from typical 2D images; all the pixels, not only the neighboring ones but also the pixels far separated, are closely related. For example, 1) multiple molecular bands such as OH*, CH*, $C_2$*, and $H_2O$* that are far separated in the spectrum will get stronger at the same time as ϕ increases, 2) the widths of the bands will broaden as P rises, and 3) the broadband background emission will become stronger as the gas density and $CO_2$ concentration increase. In short, expanding the receptive field for exploiting the global contextual information of the emission spectra would improve the prediction accuracy while keeping it below 2 × W, because the pixels in the emission spectrum are all related (unlike in common 2D images).

### 3.4. Computational efficiency

Computational efficiency is another important parameter that defines the performance of a deep learning-based method. The accuracy and precision of prediction can be improved by increasing the receptive field as shown in Fig. 10; however, the increasing $N_l$ and $N_k$ along with the receptive field costs the neural network computational complexity to lengthen the forward calculating time (running time shown in Fig. 11) and limits real-time gas property measurements. Nevertheless, increasing the receptive field via implementing the DU operators and increasing $N_d$ accelerates the running and training of a neural network model [27]. The computation time of the proposed CNN model is estimated in a Python environment on a computer with an NVIDIA GeForce RTX 3090, an AMD Ryzen 5 2600X, and 32GB of RAM as illustrated in Figs. 11 (a) and (b). The running time shown in Fig. 11 (a) is calculated by averaging the time with 5,000 inputs using batch size as 1, and the training time in Fig. 11 (b) is estimated by the time used to train a model for 100 epochs.

The trends observed in Figs. 11 (a) and (b) confirm that the DU operator accelerates the computation under a given receptive field; widening the receptive field of the CNN by increasing $N_d$ is much more cost-effective than increasing $N_k$ or $N_l$ while $N_d$, $N_l$, or $N_k$ increases the receptive field by the same order of magnitude (Eqn. (1)). The DU operator ($N_d > 1$) reduces the computational time because the range of convolution would decrease via reshaping the input pixels into sub-signals. Moreover, the number of training parameters ($N_p$), calculated as Eqn. (4), is always smaller with increased $N_d$ than that with increased $N_k$ or $N_l$ at a given receptive field as shown in Fig. 11 (c); increasing $N_p$ adds computational cost, i.e., complexity of the CNN architecture.

$$\# \ of \ paramter \ (N_p) = (N_l - 2) \times (N_c^2 N_k + N_c) + 2N_d N_c N_k + (N_c + N_d) \qquad (4)$$

$N_p$ increases with all of the three parameters, $N_l$, $N_k$, and $N_d$; however, $N_l$ increases $N_p$ by factor of $N_c$ while $N_l$ and $N_k$ increase $N_p$ by factor of $N_c^2$ ($> N_c$). Thus, $N_d$ less affects the computational cost.

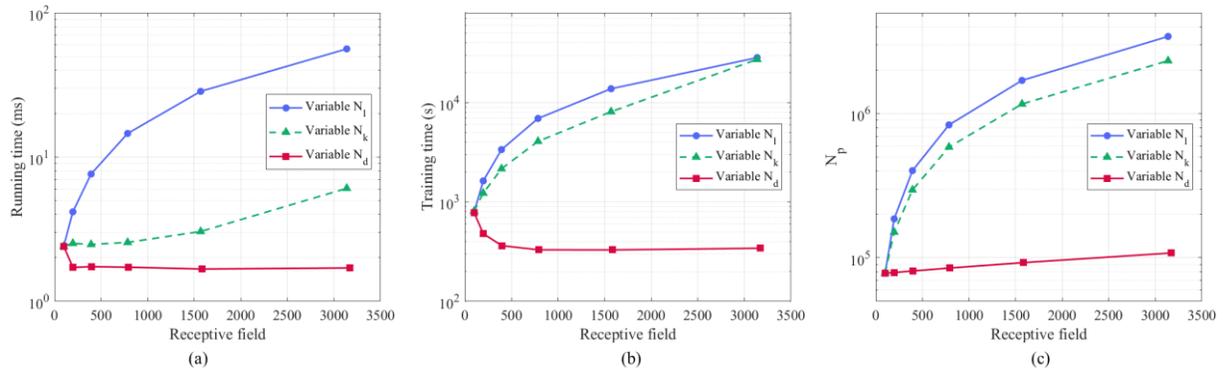

Figure 11 (a) Running time, (b) training time, and (c) $N_p$ versus receptive fields of plain CNN ($N_d = 1$) and DU + CNN / POD loss ($N_d = 2 - 32$)

## 3.5 Exposure time

To monitor the gas properties in fast-evolving combustion environments using FES, the exposure time should be matched with the characteristic time scale. For example, exposure times of several milliseconds (ms) are required to resolve large-scale eddies and precessing vortex core of swirl flames [1-4]. Furthermore, exposure times of tens and hundreds of microseconds (μs) are required to fully resolve the turbulent flame dynamics considering the integral and Kolmogorov time scale, respectively [5, 6]. In this section, the effect of the reduced exposure time on the performance of the calibration model is evaluated by repeating the experiments with varied exposure times, 0.05, 0.2, and 0.4 s. The effective minimum exposure time in this study was reduced down to 0.05 s from 2 s using the calibration model, which is in practice limited by the sensitivity of the un-intensified potable detection system used in the experiment. The minimum effective exposure time can further be reduced by a few orders of magnitude when using photon signal intensifiers. In Table 2, each component of the noise with the three different exposure times at P = 10 bar and ϕ = 1 is estimated. Presumably, the constant read-out noise is responsible for the decreased SNR with the reduced exposure time. Figure 12 (a) confirms that the SNR of the chemiluminescence signal decreases with reduced exposure time; the average SNR of OH* at 308nm is 4.3, 14, and 22 for 0.05, 0.2, and 0.4 s exposure, respectively. Prediction accuracies of LS, plain CNN, and DU + CNN / POD loss, defined as 100 minus REP, are presented in Figs. 12 (b) and (c). As expected, the accuracy drops with decreasing exposure time. Again, the DU + CNN / POD loss model performs the best (the most accurate) among the three calibration methods; P- and ϕ-prediction accuracies (100 minus REP) are above 80% and 95%, respectively, even with 0.05 s exposure.

**Table 2 Instrumental noise analysis of OH* signal at 308 nm at P = 10 bar and ϕ = 1.**

| Noise source Unit: electrons | Exposure time | | |
|---|---|---|---|
| | 0.05 s | 0.2 s | 0.4 s |
| $N_{electron}$ | 115 | 463 | 934 |
| $\sigma_{Photon}$ | 11 | 22 | 31 |
| $\sigma_{Dark}$ | 4.6 | 9.1 | 13 |
| $\sigma_{Read}$ | 24 | 24 | 24 |
| SNR | 4.3 | 14 | 22 |

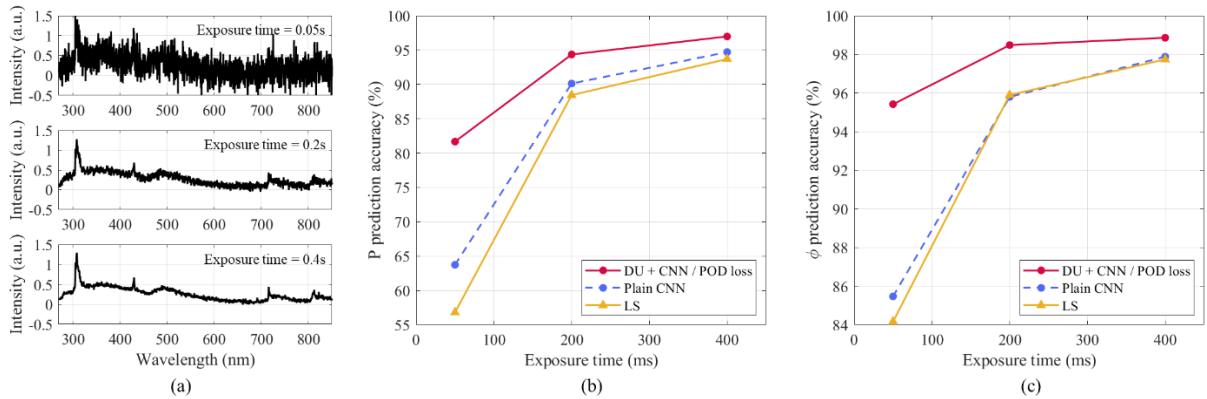

**Figure 12 (a) Typical chemiluminescence signals with the three different exposure times (0.05, 0.2, and 0.4 s) at P = 10 bar and ϕ = 1, and (b) P- and (c) ϕ-prediction accuracy versus exposure time.**

## 4. Conclusion Remarks

A calibration process employing deep-learning architecture and data mapping based on a data-driven technique was employed to enable accurate measurements of pressure (P) and equivalence ratio ($\phi$) using short-gated flame emission spectra. The deep-learning architecture utilizes a reversible down- and up-sampling (DU) operator, deep CNN layers, and is guided by loss function based on POD coefficients to train with the pairs of low-SNR (LS) and high-SNR (HS) spectra, i.e., training data, captured with short and long exposure times, respectively. The proposed neural network architecture successfully denoised the LS spectrum signals while conserving the characteristic spectral features in the spectrum. Then, the POD and a reduced order model (kriging) that are trained by the HS spectra from the training dataset were used to predict the gas properties from the denoised short-gated LS spectrum. The prediction errors of the P and $\phi$ employing the new technique were improved to approximately 5.7% and 1.5% with 0.2 s exposure time (from 11% and 4% without denoising but with the POD and kriging model), and to 18% and 5% with 0.05 s exposure time (from 43% and 16% without denoising), respectively.

It was confirmed that the combination of the proposed CNN model and the new loss function including POD coefficient loss can effectively decouple the noise and signal and selectively suppress the noise that increases with the decreasing exposure time by a data-driven approach. The proper choice of the down-sampling parameter ($N_d$) in the DU operator in CNN architecture reduced the model complexity ($N_l$ and $N_k$) and accelerated the calculation with the increased receptive field. Another advantage of the technique comes from the use of the proper orthogonal decomposition (POD) method in the denoising and calibration procedure. POD can effectively extract the property-sensitive components from the emission spectra. Therefore, the new loss function including POD loss can better guide the denoising CNN by considering the property-sensitive global features of the emission spectra.

The proposed calibration method is evidently capable of further improving the time resolution and accuracy of FES when used with more sensitive photon detecting systems, e.g., high-speed kHz-framing cameras. The transient property variations could be fully resolved if the exposure time is reduced below the characteristic fluctuation time using the fast photon detector in conjunction with the proposed denoising technique. In addition, the proposed denoising technique can also be used to denoise the flame emission spectra of other fuels and any other forms of multi-dimensional data, e.g., 2D/3D images, when reference data pairs of high SNR and low SNR signals are available.

# Acknowledgment

This work was supported by the Basic Research Funding of the Air Force Office of Scientific Research [FA2386-20-1-4054]; Korean Agency for Defense Development [16-106-501-035]; and the National Research Foundation of Korea [2021R1A2C2012697, 2021R1A4A1032023].